\documentclass[twocolumn]{article}
\usepackage{amssymb,latexsym,amsmath,array,delarray,psfig,natbib,here}
\renewenvironment{abstract}{\centerline{\large\bf  
 Abstract}\vspace{0.5ex}\begin{quote}}{\par\end{quote}\vskip 1ex}
\date{}
\title{Factorization of Language Models through Backing-Off Lattices}

\author{
Wei Wang
\\ Department of Computer Science
\\ New York University
\\ {\tt \{wei\}@cs.nyu.edu}
}

\begin{document}
\maketitle

\begin{abstract}
Factorization of statistical language models is the task that 
we resolve the most discriminative model into factored models
and determine a new model by combining them so as to 
provide better estimate.
Most of previous works mainly focus on factorizing
models of {\it sequential} events, each of which 
allows only one factorization manner. To enable {\it parallel}
factorization, which allows a model event to be resolved 
in more than one ways at the same time, we propose a general
framework, where we adopt a backing-off lattice to
reflect parallel factorizations and to define
the paths along which a model is resolved into 
factored models, we use  a mixture model to combine
parallel paths in the lattice, and generalize
Katz's backing-off method to integrate all
the mixture models got by traversing the entire
lattice.
Based on this framework, we formulate
two types of model factorizations that are used in 
natural language modeling.
\end{abstract}

\section{Introduction}
Factorization of statistical language models is the task that 
we resolve the most discriminative model into factored models
and determine a new model by combining them so as to 
provide better estimate to the most discriminative model event.
For instance,
a new model for trigram can be obtained by combining the factored
models: a unigram model, a bigram model and a trigram model; a model
for PP-attachment \citep{collins-brooks-backoff} can be obtained by
considering both
more
discriminative models like $\Pr(1|is, revenue, from, research)$
\footnote{
This example is extracted from \citep{collins-brooks-backoff}.
}
and less discriminative ones like $\Pr(1|is, from, research)$;
a lexicalized parsing model can be approximated by combining 
a lexical dependency model and a syntactic
structure model \citep{klein-manning-factor}. The former two examples are
usually called backing-off.

Therefore, factorization of language models should answer
two questions: how to factorize, and how to combine.
Most of previous works on language modeling \citep{chen-goodman} 
\citep{goodman}
focus on sequential model event (such as n-gram), and thus 
need not to answer the first question because the sequential
model event like n-gram gives a natural factorization order:
an n-gram has exactly one type of (n-1)-gram to backoff. 
However, for nonsequential model event, we need to specify
them both.

In this paper, we formulated a framework for language model
factorization.  We adopt a backing-off lattice to
reflect parallel factorization and to define
the paths along which a model is resolved into 
factored models; we use  a mixture model to combine
parallel paths in the lattice; and generalize
Katz's backing-off method to integrate all
the mixture models got by traversing the entire
lattice.

Based on this framework, we formulate
two types of model factorizations that are used in 
natural language modeling.

The remainder of this paper is organized as follows,
we first introduce the backing-off lattice,
 then explain the mixture model, next formulate the backing-off
formula, next describe two
types of model factorizations, 
and finally draw the conclusions.

\section{Backing-Off Lattice}
\label{sec:lattice}
The backing-off lattice specifies the ways how an event can 
be ``factorized'' into sub-events. Each lattice node represents
a {\it set} of factored events\footnote{We also regard the
most discriminative events in a model as factored events.}.
Each lattice edge connects a parent node to a child node, 
and represents a factorization manner that factorizes
an event in the parent event set into a set of factorized
events represented by its child node. Different lattice nodes may
have common child.

In most of previous works, the backing-off lattice is only
a list, in which no node has more than one edges (backing-off paths). 
Our backing-off lattice is, however, an directed acyclic graph
(DAG), which means a model event 
represented by a lattice node might have several 
factorization manners.

\begin{figure}
\hspace{1cm}
\psfig{figure=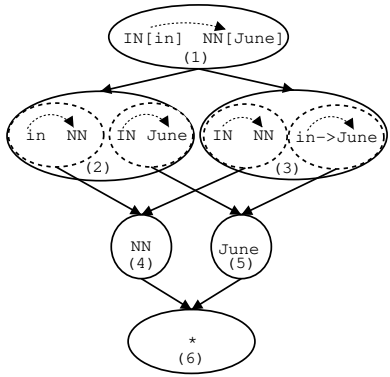,width=2in}
\caption{
\footnotesize
A backing-off lattice for factorization of a dependency event 
in a bilexical context-free grammar. Each lattice node 
is denoted by a solid oval and represents a set of events,
each of which is represented
by a dotted oval. Each lattice edge represents a factorization 
manner that resolves
an event in the parent node (e.g., the left dotted
oval in node (3)) into a set of factored events
(e.g., the set of events in node (4)).
\label{backoff-lattice}}
\end{figure}

Figure~\ref{backoff-lattice} shows a backing-off lattice
that illustrates how to factorize a dependency event
in a bilexical context-free grammar \citep{es99}.
Each lattice node is denoted by a solid oval and 
represents a set of events, each of which is represented
by a dotted oval (if there is only one element in the set,
we omit the dotted oval).  
Each lattice edge represents a factorization 
manner that resolves
an event in the parent node (e.g., the left dotted
oval in node (3)) into a set of factored events
(e.g., the set of events in node (4)).

The backing-off lattice should be tailored in accordance
with the requirement of the task that it is applied to.
If it is used for smoothing purpose, we may want
to use each slice of the lattice to represent
model events with the same specificity, which is 
less than its previous slice. To combine different
resources, we may want to factorize a complex model
whose statistics are unavailable to factored models
whose statistics are available.

\section{Mixture Model}
\label{sec:mixture}
Through the backing-off lattice, a model is factorized
recursively into sub-models. Each node can be applied
with more than one factorization manners.
We therefore are concerned with the problem
of how to approximate the model 
represented by a lattice node
through the models represented by
all its children nodes. For instance,
how to approximate the distribution
of events in node 1 of Figure~\ref{backoff-lattice}
with factored models represented by nodes 2 and 3.
We use a mixture
model to interpolates all the factored
models. We formulate the mixture model in the following.

Let $\mathcal{E}$ denote a backed-off event (e.g., the Dependency event in 
the lattice root in figure \ref{backoff-lattice}).
To get its factored events, we introduce a series of
{\it factorization function}s: $\Phi_{i}, 0 \le i \le I$, each of
which 
corresponds to a lattice edge from $\mathcal{E}$, and
factorize $\mathcal{E}$ into a set of sub model events $\mathcal{S}_{i}$,
that is, 
\begin{eqnarray}
\mathcal{E} 
\begin{array}{c}
\Phi_{i} \\
\longrightarrow \\
   \\
\end{array}
\mathcal{S}_{i} = \{e_{ij} | 0 \le j \le I_{i}\}
\label{project-function}
\end{eqnarray}

where, $e_{ij}$ is the $j$'th sub-event among the set of events obtained
by factorizing $\mathcal{E}$ using factorization function $\Phi_{i}$. 

We can view $\Phi_{i}$ as a hidden random variable, corresponding to 
different a factorization manner. It has a prior distribution:
$\Pr(\Phi_{i})$, specifying the confidence of selecting the $i$'th path to 
backoff.  We can get,
\begin{eqnarray}
\label{2}
\Pr(\mathcal{S}_{i}) = \Pr(\mathcal{E}| \Phi_{i})
\end{eqnarray}
The distribution of $\mathcal{E}$ can be derived in the following way,
\begin{eqnarray}
\label{mixture-formula}
\Pr(\mathcal{E})& = & \sum_{\Phi_{i}, 0 \le i \le I}\Pr(\mathcal{E}, \Phi_{i})
                           \nonumber \\
                 & =&  \sum_{\Phi_{i}, 0 \le i \le I}\Pr(\Phi_{i})
		                                 \Pr(\mathcal{E}|\Phi_{i})
\end{eqnarray}
From Formula~\ref{2} and \ref{mixture-formula}, we get
\begin{eqnarray}
\label{mixture-of-factored-model}
\Pr(\mathcal{E}) = \sum_{\Phi_{i}, 0 \le i \le I}\Pr(\Phi_{i})\Pr(\mathcal{S}_{i})
\end{eqnarray}

Formula~\ref{mixture-of-factored-model} shows that the probabilistic
model governing event $\mathcal{E}$ is approximated by a
mixture of its factored models $\Pr(\mathcal{S}_{i})$ 
using normalized coefficients ($\Pr(\Phi_{i})$). Each 
coefficient reflects the confidence of selecting a factorization
manner.

The sum of $\Phi_{i}$ should be equal to 1. Their values can be handcrafted
just for simplicity or trained from
held-out data using EM algorithm \citep{em} or other numerical
methods such as Powell's method \citep{powell}.

If we assume that the factored events $s_{ij}$ in the value
set of a factorization function are independent of each other,
 we arrive
\begin{eqnarray}
\label{independent}
\Pr(\mathcal{E}) = \sum_{\Phi_{i}, 0 \le i \le I}\Pr(\Phi_{i})
		                                 \prod_{j}^{|\mathcal{S}_{i}|}\Pr(e_{ij})
\end{eqnarray}

Formula~\ref{independent} shows that if we assume 
the events in the value set of each projection
function are independent of each other, the probability
of the value set given factorization function is equal to
the multiplication of the probability of each event in the set.

We can derive the mixture model for conditional distributions
similarly.

Let us give an example to illustrates the above idea. In the 
backing-off lattice shown in Figure~\ref{backoff-lattice},
for the event in node 1 to be factored into event sets
in node 2 and 3, respectively, we need to introduce the
following factorization functions,

\begin{enumerate}
\item[]$\Phi_{0}$=   {\it factorize $\mathcal{E}$ into a
                      lexical dependency and a syntactic dependency.}

\item[]$\Phi_{1}$= {\it factorize $\mathcal{E}$ into 
                sub-events, each of which describes
		the dependency between (parent or dependent) lexical head and 
		(dependent or parent) nonterminal label.}
\end{enumerate}

These functions will project the event into
\begin{eqnarray}
\mathcal{S}_{0} = \{in\rightarrow June, IN \rightarrow NN\}. \\
\mathcal{S}_{1} = \{in\rightarrow NN, IN \rightarrow June\}.
\end{eqnarray}
where each of the two sets contains two factored dependency
events.

Then based on Formula~\ref{mixture-of-factored-model}
we get
\begin{eqnarray}
\Pr(IN[in]\rightarrow NN[June]) \hspace{2cm}  \nonumber \\
= \Pr(\Phi_{0})\Pr(in \rightarrow June, IN \rightarrow NN)   \nonumber\\
+ \Pr(\Phi_{1})\Pr(in \rightarrow NN, IN \rightarrow June) 
\end{eqnarray}

Assume that factored events in the same set are independent 
of each other, from Formula~\ref{independent}, we can get
\begin{eqnarray}
\Pr(IN[in]\rightarrow NN[June]) \hspace{2.5cm}  \nonumber \\
= \Pr(\Phi_{0})\Pr(in \rightarrow June)\Pr(IN \rightarrow NN) \nonumber\\
+ \Pr(\Phi_{1})\Pr(in \rightarrow NN)\Pr(IN \rightarrow June) 
\end{eqnarray}

Now, distribution of $\Pr(IN[in] \rightarrow NN[June])$ is 
approximated by the mixture of the factored models 
$\Pr(in \rightarrow June)$, $\Pr(IN \rightarrow NN)$ , 
$\Pr(in \rightarrow NN)$ and $\Pr(IN \rightarrow June)$ that
are obtained based on a pre-defined  backing-off lattice.

\section{Backing-Off Formula}
\label{sec:formula}
We have presented a mixture model to approximate
a more discriminative model with
less discriminative factored models
based on a backing-off lattice and a set of 
factorization functions. A mixture model, however,
only combines the factored models obtained
by factorizing one lattice node and if we traverse the backing-off
lattice, we will get a series of mixtures.
We therefore 
generalize Katz's backing-off method \citep{katz}
to organize these mixtures 
by firing correspondent mixture model 
when backing-off takes place.

\begin{eqnarray}
\Pr_{bo}(\mathcal{E}_{1}|\mathcal{E}_{2}) = \hspace{5cm} \nonumber \\
  \left\{ \begin{array}{ll}
   \frac{C(\mathcal{E}_{1}\mathcal{E}_{2})}{C(\mathcal{E}_{2})} &  C(\mathcal{E}_{1}\mathcal{E}_{2}) > K       \\
   \beta_{C(\mathcal{E}_{1}\mathcal{E}_{2})}\frac{C(\mathcal{E}_{1}\mathcal{E}_{2})}{C(\mathcal{E}_{2})} &  1 \le C(\mathcal{E}_{1}\mathcal{E}_{2}) \le K \\
   \alpha \mathcal{MIXTURE} & otherwise
   \end{array} \right.
\label{backoff-formula}
\end{eqnarray}
  
where, $\mathcal{MIXTURE}$ represents the conditional
version of the mixture model in Formula~\ref{mixture-of-factored-model}.
$K$ is a frequency threshold for discounting. $\mathcal{E}_{1}$ and $\mathcal{E}_{2}$
refer to, in general, two events that adjacently co-occur in a corpus.

The basic idea of this backing-off
method is the same as that of Katz's. 
That is, the backing-off formula has
a recursive format. At each step of the recursion, there are three branches 
associated with their firing conditions. If the frequency of the current
model event is large enough (such as greater than $K$, Katz used the value
of 5 for $K$), the 
maximum-likelihood estimator (MLE) is used. If the occurrence frequency is
within the range of $[1, K]$, the MLE probabilities are discounted in some
manner so that 
some probability mass is reserved for those unseen events. If the model event
never occurs in the training data, we use the estimates from the factored 
model events. 

The difference therefore lies in the combination of estimates of factored 
events. In traditional backing-off methods, there is only one backing-off path
to go when the backing-off condition is satisfied. For example, an n-gram only
has exactly one (n-1)-gram to be backed-off. However, in our case, we have 
more than
one backing-off paths to go through. None is a branch of another. 
Then the mixture model obtained in the previous section 
is embedded here.

$\beta$ are for normalization and can be computed according
to \citep{katz}.

$\alpha$ is also for normalization. It is
 computed from the amount of reserved probability mass for unseen events.
$\alpha$ is a function of $\mathcal{E}_{2}$ because $\mathcal{E}_{2}$ is the given event of
a conditional probability, and each conditional probability should satisfy
the normalization requirement.  
It is computed similarly to that in Katz original paper: 

\begin{eqnarray}
\alpha & = &\alpha(\mathcal{E}_{2}) \nonumber \\
 & = & \frac{1-\sum_{\mathcal{E}_{1},\mathcal{E}_{2}:C(\mathcal{E}_{1}\mathcal{E}_{2}) > 0} \Pr_{bo}(\mathcal{E}_{1}|\mathcal{E}_{2})}
                   {\sum_{\mathcal{E}_{1}, \mathcal{E}_{2}:C(\mathcal{E}_{1}\mathcal{E}_{2})=0}\mathcal{MIXTURE}} \nonumber  \\
       &  = &\frac{1-\sum_{\mathcal{E}_{1},\mathcal{E}_{2}:C(\mathcal{E}_{1}\mathcal{E}_{2}) > 0}\Pr_{bo}(\mathcal{E}_{1}|\mathcal{E}_{2})}
                           {1-\sum_{\mathcal{E}_{1}, \mathcal{E}_{2}:C(\mathcal{E}_{1}\mathcal{E}_{2})>0}\mathcal{MIXTURE}}
\label{bow-formula}
\end{eqnarray}

\section{Model Factorizations}
\label{sec:types}
Now that we have presented a framework that allows a 
model event to be factorized along more than paths and combines
different paths in a backing-off formula, we now formulate
two types of model factorization that are used in 
natural language modeling. We first introduce some
notations.

Let a matrix $\mathcal{M}^{m \times n}$ of random variables
represent a linguistic object that simultaneously expresses
two types of information in its row and column directions.
For example, matrix\footnote{ This example is due to the hierarchical
alignment in Figure 10 in \citep{alshawi-fst}},
\begin{eqnarray}
\label{alshawi-example}
\left[\begin{array}{ll}
nonstop & flights \\
sin     & vuelos 
\end{array} \right]
\end{eqnarray}

denotes two lexical dependencies, each of which
is the translation the other. It expresses
the dependency relationship information in the 
row direction, and translation relationship
information in the column direction.

Let $\mathcal{A} \in \mathcal{M}^{m \times n}$,
and $\mathcal{B}$\footnote{
     We let $\mathcal{A}$ and $\mathcal{B}$ have 
     identical distributions for simplicity
     of explanation.  They needn't have to be identical
     in general.
     And matrix $\mathcal{M}$ need not to have only
     row and column, but might be like $\mathcal{M}^{m \times n \times l \dots}$
     .
}
$\in \mathcal{M}^{m \times n}$,
we want to determine $\Pr(\mathcal{A} | \mathcal{B})$
using its factored models. We can either factorize
both $\mathcal{A}$ and $\mathcal{B}$ synchronously,
or factorize only the conditioning event $\mathcal{B}$,
which results in two types of factorizations for
different tasks: {\it synchronous factorization},
and {\it asynchronous factorization}.

\subsection{Synchronous factorization}
In synchronous factorization, both $\mathcal{A}$ 
and $\mathcal{B}$ are factorized in the same manner
according to some correspondence assumption,
and the factored models determines the marginal 
information of the entire model.

Based on Formula~\ref{mixture-formula} (the mixture
model), and the assumption that
$\mathcal{A}$ and $\mathcal{B}$ are in sync with 
with each other on row (or column),
we formulate the synchronous
factorization as follows:

\begin{eqnarray}
\label{synchronous-factorization}
\Pr(\mathcal{A}|\mathcal{B}) &= & 
 \Pr\left(\Phi_{row}\right)\Pr\left(\mathcal{S}_{row}^{\mathcal{A}}|
                    \mathcal{S}_{row}^{\mathcal{B}}\right)  \nonumber \\
 &+& \Pr\left(\Phi_{col}\;\;\right)\Pr\left(\mathcal{S}_{col}^{\mathcal{A}}\;|
                    \mathcal{S}_{col}^{\mathcal{B}}\;\right)
\end{eqnarray}

where,
\begin{itemize}
\item[-] $\Phi_{row}$ projects $\mathcal{A}$ and $\mathcal{B}$ on row,
                    respectively, and therefore results in
		    row vectors:
          \begin{enumerate}
	    \item[-] $\mathcal{S}_{row}^{\mathcal{A}}$= 
		    $\{(\mathcal{A}_{ij})_{1 \times \ n} | 1 \le i \le m\}$

		    the set of row vectors of $\mathcal{A}$
	    \item[-] $\mathcal{S}_{row}^{\mathcal{B}}$= 
		    $\{(\mathcal{B}_{ij})_{1 \times n} \;\;| 1 \le i \le m\}$
		    
		    the set of row vectors of $\mathcal{B}$
	  \end{enumerate}
\item[-] $\Phi_{col}$ projects $\mathcal{A}$ and $\mathcal{B}$ on column,
                      respectively, and therefore results in
		      column vectors:
           \begin{enumerate}
	    \item[-] $\mathcal{S}_{col}^{\mathcal{A}}\;$= 
		    $\{(\mathcal{A}_{ij})_{m \times 1} \;| 1 \le j \le n\}$

		    the set of column vectors of $\mathcal{A}$
	    \item[-] $\mathcal{S}_{col}^{\mathcal{B}}\;$= 
		    $\{(\mathcal{B}_{ij})_{m \times 1} \;\;| 1 \le j \le n\}$

		    the set of column vectors of $\mathcal{B}$
           \end{enumerate}
\end{itemize}

Factored models 
$\Pr\left(\mathcal{S}_{row}^{\mathcal{A}}|\mathcal{S}_{row}^{\mathcal{B}}\right)$
and 
$\Pr\left(\mathcal{S}_{col}^{\mathcal{A}}|\mathcal{S}_{col}^{\mathcal{B}}\right)$
will be further factorized by other factorization functions, and the
results of these factorization functions constitute the backing-off lattice.
All these factored models are then combined by the backing-off 
formula.

Let us give an example. Let 
\begin{eqnarray}
\label{depend}
\mathcal{A} = \left[\begin{array}{l} nonstop \\ sin \end{array}\right]
\end{eqnarray}
and let 
\begin{eqnarray}
\label{parent}
\mathcal{B} = \left[\begin{array}{l} flights\\ vuelos\end{array}\right]
\end{eqnarray}

Under factorization manner $\Phi_{row}$, and from 
Formula~\ref{synchronous-factorization}, \ref{depend} and \ref{parent}
we get
\begin{eqnarray}
\label{17}
& &\Pr\left(\begin{array}{c|c}
              \left[\begin{array}{c}nonstop \\ sin\end{array}\right] &
	      \left[\begin{array}{c}flights \\  vuelos\end{array}\right] \\
       \end{array}\right) \nonumber \\
& = & \Pr(\Phi_{row}) \Pr\left(
                 \mathcal{S}_{row}^{\mathcal{A}}|\mathcal{S}_{row}^{\mathcal{B}}
	                 \right) 
\end{eqnarray}
where
\begin{eqnarray}
\mathcal{S}_{row}^{\mathcal{A}} &=& \Phi_{row}(\mathcal{A}) 
                                = \left\{
				    \left[nonstop\right],\left[sin\right]
                                  \right\} \\
\mathcal{S}_{row}^{\mathcal{B}} &=& \Phi_{row}(\mathcal{B}) 
                                = \left\{
				    \left[flights\right],\left[vuelos\right]
                                  \right\}
\end{eqnarray}
and $\Phi_{row}=1$ (normalized).

If we assume that an element in $\mathcal{S}_{row}^{\mathcal{A}}$
is only dependent on the correspondent element in 
$\mathcal{S}_{row}^{\mathcal{B}}$, we get,

\begin{eqnarray}
\label{20}
\Pr\left(
          \mathcal{S}_{row}^{\mathcal{A}}|\mathcal{S}_{row}^{\mathcal{B}}
   \right)& =& \Pr\left(nonstop|flights\right) \nonumber \\
	   &\times & \Pr\left(sin|vuelos\right) 
\end{eqnarray}

Formula~\ref{17} and \ref{20} indicate that,
by synchronous factorization, a bilingual
lexical dependency model like that in \citep{alshawi-fst} is approximated by two factored
lexical dependency models , each of which corresponds to
one language.

The factorization of a bilexical context-free grammar into
a lexical dependency model and a syntactic structure model
in \citep{klein-manning-factor} is actually synchronous factorization.

Synchronous factorization is usually used for information
combination in the cases that we only have the statistics
of those factored models and want to use them to 
approximate a more complex model; or that we want
to simplify a complex model into factored models
to gain efficiency \citep[e.g.,\ ][]{klein-manning-factor}.

\subsection{Asynchronous factorization}
Another type of factorization that is frequently
used in statistical language modeling is asynchronous 
factorization, where only the conditioning event of
the conditional probability $\Pr(\mathcal{A}|\mathcal{B})$
is recursively factorized while keeping the conditioned
event fixed. The following formula describes the idea.

\begin{eqnarray}
\label{asynchronous}
\Pr(\mathcal{A}|\mathcal{B}) =
 \sum_{i=1}^{|\mathcal{B}|}\left(\Pr(\Phi_{i})\Pr(\mathcal{A}|\Phi_{i}(\mathcal{B}))\right)
\end{eqnarray}

where $\Phi_{i}$ = ``drop the $i$'th element in matrix $\mathcal{B}$'',
and $|\mathcal{B}|$ is the number of the elements in matrix $\mathcal{B}$.
If we further factorize model 
 $\Pr(\mathcal{A}|\Phi_{i}(\mathcal{B}))$,
matrix $\mathcal{B}$ will be a partial matrix that contains a part
of elements of the original matrix. 

Formula~\ref{asynchronous} indicates
that the original model is recursively factorized into sub-models, and
each factorization recursion step has $|\mathcal{B}|$ factorization
manners, each of which only drops one element from $\mathcal{B}$.

In the following, we give an example to illustrate 
the idea of asynchronous factorization. 
In PP-attachment, suppose we want to factorize model
$\Pr(0|is, revenue, from, research)$\footnote{
Once again, this example is extracted from \citep{collins-brooks-backoff}.
}, which
determines the probability of the attachment of preposition phrase
``from research'' to the noun ``revenue'' instead of to the verb
``is''. Based on Formula~\ref{asynchronous}, we can get,

\begin{eqnarray}
\Pr(1|is, revenue, from, research) = \nonumber \\
            \Phi_{revenue}\Pr(1|is, from, research)  \nonumber \hspace{0.2cm}\\
	    + \Phi_{research}\Pr(1|is, revenue, from) \nonumber \hspace{0.2cm} \\
	     + \Phi_{is}\Pr(1|revenue, from, research) 
\end{eqnarray}
where $\Phi_{word}$ refers to the factorization function that drops $word$ from
the conditioning event of the left hand side model.
And the factored models on the right hand side can be 
further factorized by continuing to traversing 
the backing-off lattice.

In contrast to that synchronous factorization is used
for information combination, asynchronous factorization
is usually used for smoothing purpose.

In practice, there might be a compromise between the 
above two, where we factorize both the conditioned
and conditioning events, but not in a synchronous manner. 
And matrices $\mathcal{A}$ and $\mathcal{B}$ need not
to have the same number of rows and columns.

\section{Related Works}
\citep{collins-brooks-backoff} puts forward a method,
which actually is asynchronous factorization, for backing-off
of models of prepositional phrase attachment, providing a way
to mixing the frequencies of different backing off choices at
certain recursion step by dividing the sum of frequencies of 
all the more discriminative model events by those of less
discriminative ones if the sum of the frequencies of the less
discriminative ones are greater than zero, otherwise, backing-off
continues on.  One characteristic of this method is that if one of
the less discriminative model events has non-zero frequency, the 
backing-off terminates, no matter whether other events in the 
same backing-off level are zero or not, whereas the mixture model 
we introduced to combine parallel backing-off paths is able to
to make those zero count event further backoff so 
that they still can contribute to the final result. And we think
this is necessary when we use the backing-off framework for
information combination.

It turned out that \citep{parallel-backoff} were independently
working on some similar ideas. They introduced factored models
and also generalized the backing-off framework to handle parallel
backing-off paths. The differences between their work and ours
are (1) Their factored models can actually be categorized into
the asynchronous factorization type, where only the conditioning
matrix (the feature vector in their paper) is factorized;
(2) We also formulated the synchronous factorization type, where
both the conditioned and conditioning matrices are factorized
synchronously. And we showed that this is useful for combining
different information sources; (3) We use a mixture model to
combine parallel paths while they selected the the path with
the maximum value. We think that combining the contribution
of each backing-off path is useful when we want to combine
different information resources; (3) In our framework, the result
of each factorization function (backing-off path) is a set of
events (See Formula~\ref{project-function}),
not merely one event. And this is usefull when we
do the kind of factorization like \citep{klein-manning-factor}
using Formula~\ref{independent}.

\section{Conclusions}
We have presented a framework for language model
factorization.  We adopt a backing-off lattice to
reflect parallel factorizations and to define
the paths along which a model is resolved into 
factored models, we use  a mixture model to combine
parallel paths in the lattice, and generalize
Katz's backing-off method to integrate all
the mixture models got by traversing the entire
lattice.

Based on this framework, we formulate
two types of model factorizations that are used in 
natural language modeling.

\end{document}